\documentclass[lettersize,journal]{IEEEtran}
\usepackage{amsmath,amssymb,amsfonts}
\usepackage{algorithmicx,algorithm}
\usepackage{amsthm,amsmath,amssymb}
\usepackage{mathrsfs}
\usepackage{array}
\usepackage{textcomp}
\usepackage{stfloats}
\usepackage{verbatim}
\usepackage{graphics} 
\usepackage{epsfig} 
\usepackage{times} 
\usepackage{amsmath} 
\usepackage{amssymb}  
\usepackage{cite} 
\usepackage{graphicx}
\usepackage{subfigure} 
\usepackage{picinpar}
\usepackage{url}
\usepackage{flushend}
\usepackage{colortbl}
\usepackage{soul}
\usepackage{multirow}
\usepackage{pifont}
\usepackage{color}
\usepackage{alltt}
\usepackage{hyperref}
\usepackage{float}
\usepackage{enumerate}
\usepackage{siunitx}
\usepackage{breakurl}
\usepackage{epstopdf}
\usepackage{pbox}
\usepackage{booktabs}
\usepackage{makecell} 
\usepackage{threeparttable} 
\usepackage{balance}
\usepackage[justification=centering,labelsep=period]{caption} 
\usepackage{color} 
\usepackage{gensymb} 
\usepackage{upgreek}
\hypersetup{colorlinks=true, linkcolor=black} 

\usepackage{caption}
\usepackage{adjustbox}
\usepackage{xcolor}
\usepackage{soul}
\soulregister\cite7 
\soulregister\citep7 
\soulregister\citet7 
\soulregister\ref7 
\soulregister\pageref7 

\usepackage{bm}
\usepackage{geometry}
\title{
    UrbanV2X: A Multisensory Vehicle-Infrastructure Dataset for Cooperative Navigation in Urban Areas
}

\author{Qijun Qin$^{1*}$, Ziqi Zhang$^{1*}$, Yihan~Zhong$^{1*}$,  Feng~Huang$^{1*}$,  Xikun Liu$^{1}$, Runzhi Hu$^{1}$, Hang Chen$^{2}$, Wei Hu$^{2}$, Dongzhe Su$^{2}$, Jun Zhang$^{3}$, Hoi-Fung Ng$^{1}$, Weisong~Wen$^{1 \text{\textdagger}}$     

    \thanks{
         This work was supported in part by the Innovation and Technology Fund under the project "Safety-Certified Multi-Source Fusion Positioning for Autonomous Vehicles in Complex Scenarios (ZPE8)", in part by Innovation and Technology Fund under project "Advanced Smart Mobility Road-Side and Edge System (ART/369CP)".}%
        
    \thanks{
        *Equal contribution; 
         $^{\text{\textdagger}}$Corresponding Author.
         
        $^{\text{1}}$Qijun Qin, Ziqi Zhang, Feng Huang, Yihang Zhong, Xikun Liu, Runzhi Hu, Hoi-Fung Ng and Weisong Wen are with the Department of Aeronautical and Aviation Engineering, The Hong Kong Polytechnic University, Hong Kong SAR, China. 
        
        $^{\text{2}}$Hang Chen, Wei Hu and Dongzhe Su are with the Hong Kong Applied Science and Technology Research Institute (ASTRI), Hong Kong. (e-mail: kevinchen@astri.org,  , dzsu@astri.org).

        $^{\text{3}}$Jun Zhang is with the Nanyang Technological University, Singapore. (e-mail: jzhang061@e.ntu.edu.sg). 
        }%

}

\begin{document}   
\maketitle  
\thispagestyle{headings} 
\pagestyle{headings}

\begin{abstract}
Due to the limitations of a single autonomous vehicle, Cellular Vehicle-to-Everything (C-V2X) technology opens a new window for achieving fully autonomous driving through sensor information sharing. However, real-world datasets supporting vehicle–infrastructure cooperative navigation in complex urban environments remain rare. To address this gap, we present UrbanV2X, a comprehensive multisensory dataset collected from vehicles and roadside infrastructure in the Hong Kong C-V2X testbed, designed to support research on smart mobility applications in dense urban areas. Our onboard platform provides synchronized data from multiple industrial cameras, LiDARs, 4D radar, Ultra-wideband (UWB), IMU, and high-precision GNSS-RTK/INS navigation systems. Meanwhile, our roadside infrastructure provides LiDAR, GNSS, and UWB measurements. The entire vehicle–infrastructure platform is synchronized using the Precision Time Protocol (PTP), with sensor calibration data provided. We also benchmark various navigation algorithms to evaluate the collected cooperative data. The dataset is publicly available at: \url{https://polyu-taslab.github.io/UrbanV2X/}.
%
\end{abstract}


\section{Introduction}
\label{Introduction}

\begin{figure}[htb]  
        \centering
        \captionsetup{justification=justified}
        \includegraphics[width=1.0\columnwidth]{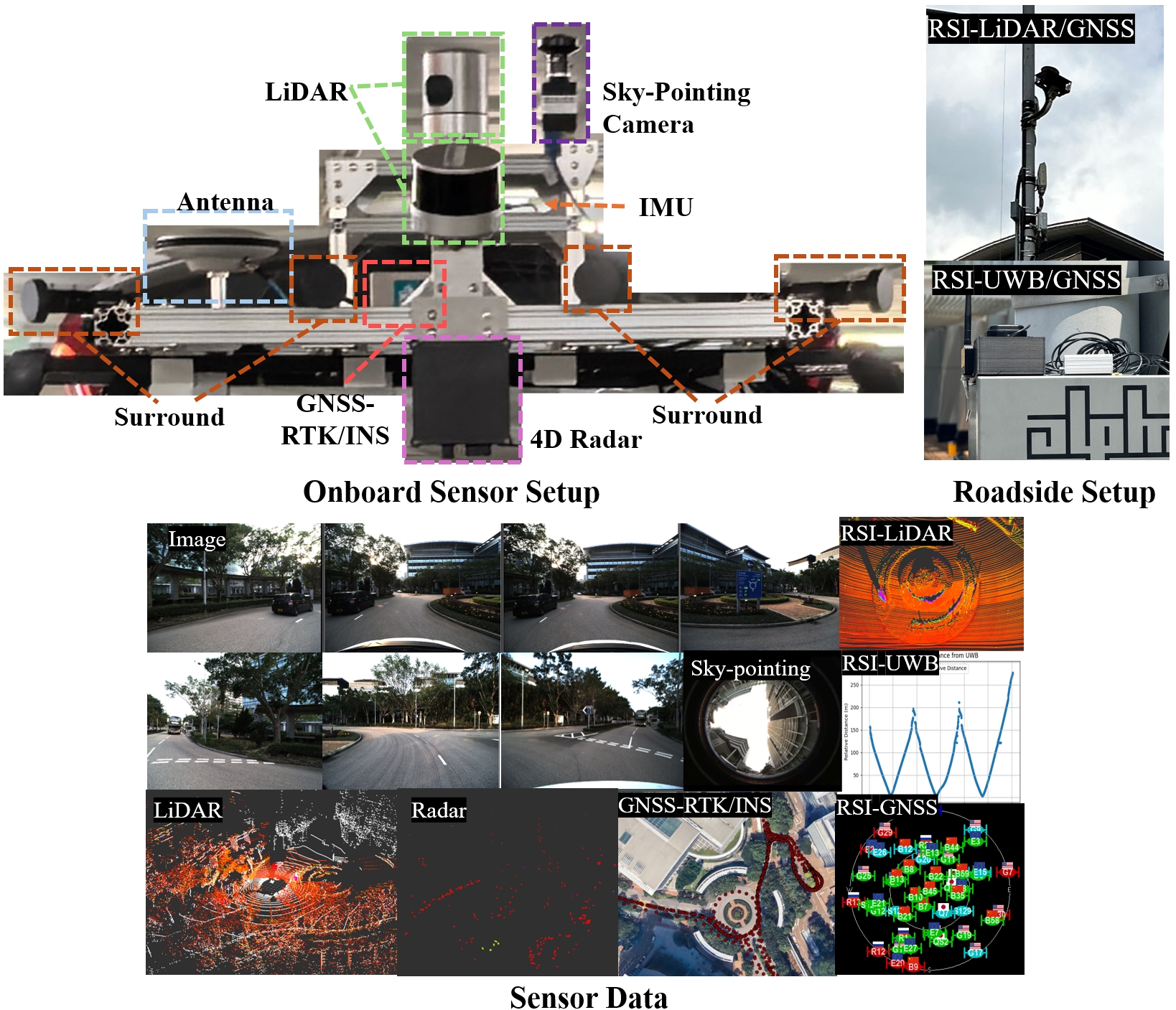}
        \caption{An overview of the setup of vehicle sensor-kit and roadside equipment, with visualization of the dataset.  Top: the vehicle sensor-kit and roadside sensors setup.  Each sensor is marked with a corresponding labeled box.  Bottom: the visualization dataset, including multiple surround-view images, a sky-pointing image, point clouds from LiDAR and 4D radar, GNSS-RTK/INS and UWB data.}  
        \label{Fig_1_sensor_setup}
        \vspace{-2.0em}
\end{figure}%
\IEEEPARstart{R}{ecent} advancements of multi-sensor integration \cite{8916930} for autonomous vehicles (AVs) highlight the significant potential of smart mobility solutions. However, positioning performance in urban areas can be severely degraded due to LiDAR degeneration caused by dynamic objects\cite{hf2022benchmark}, insufficient visual features \cite{Bai2021DegeneratedVIO} for camera-based methods, and GNSS errors stemming from multipath effects and non-line-of-sight (NLOS) conditions\cite{hsu2018analysis}. To enhance the robustness of AV systems, Cellular Vehicle-to-Everything (C-V2X) technology opens new opportunities for fully autonomous driving by enabling real-time sensor information sharing between vehicles and roadside infrastructure (RSI). According to the Connected Autonomous Vehicle (CAV) Development Study \cite{deloitte_astri_2024} conducted by Deloitte China and Hong Kong ASTRI in April 2024, roadside infrastructure and sensors are fundamental components in the CAV ecosystems. 

With the advancement of 5G technology and roadside infrastructures (RSIs), vehicle--infrastructure collaboration has gained significant attention for its potential in intelligent transportation systems~\cite{yu2022dairv2x,Zhong2024ITSC}. However, real-world datasets that support research and benchmarking in this area remain scarce. Most existing datasets focus on single-vehicle sensor configurations~\cite{KITTI,Caesar2019-fh}, simulated environments~\cite{Li9835036}, or oversimplified urban scenarios~\cite{zimmer2024tumtrafv2x}. The lack of comprehensive data from both vehicles and infrastructure in complex urban settings presents a major obstacle to advancing cooperative autonomy technologies.

To address this gap, we introduce UrbanV2X, a comprehensive and publicly available multisensory dataset designed for vehicle–infrastructure cooperative navigation in dense urban areas. Collected from the Hong Kong C-V2X testbed, UrbanV2X provides time-synchronized and calibrated data from a wide range of onboard and roadside sensors. The onboard platform includes multiple industrial cameras, LiDARs, a 4D radar, Ultra-wideband (UWB) modules, an IMU, and a high-precision GNSS-RTK/INS system. Complementing this, the roadside infrastructure is equipped with LiDARs, GNSS receivers, and UWB anchors, enabling cross-view cooperative navigation.

The contributions of our work can be summarized as follows:
\begin{enumerate}

\item 
We present UrbanV2X, a comprehensive, real-world vehicle–infrastructure dataset collected in Hong Kong's dense urban environments, as illustrated in Fig. \ref{Fig_1_sensor_setup}. It includes time-synchronized and calibrated data from a rich suite of onboard and roadside sensors (cameras, LiDARs, 4D radar, GNSS-RTK/INS, UWB, IMU), specifically designed to support research in cooperative navigation, SLAM, and smart mobility applications. UrbanV2X enables robust localization and precise positioning by integrating multimodal data from vehicles and infrastructure, addressing challenges like GNSS multipath, NLOS conditions, and dynamic urban scenarios. It supports a wide range of applications, including cooperative navigation, cross-view perception, and intelligent transportation system development, providing a valuable benchmark for advancing V2X technologies and autonomous driving.


\item 
We evaluate multiple state-of-the-art positioning algorithms using the UrbanV2X dataset, demonstrating the value of cooperative navigation.
Moreover, our dataset and benchmark results are released publicly available on our \href{https://polyu-taslab.github.io/UrbanV2X/}{website}.


\end{enumerate}

The remainder of the paper is organized as follows:
Section \ref{Related works} introduces the related works. 
Section \ref{System Overview} presents the sensor setup and sensor calibration.
Section \ref{Dataset Overview} introduces the dataset overview.
Section \ref{Dataset Evaluation} demonstrates the dataset evaluation results.
The conclusion is given in Section \ref{CONCLUSIONS}.

\section{Related works}
\label{Related works}
The design and diversity of multisensory datasets have played a crucial role in advancing perception and localization algorithms in autonomous driving and cooperative vehicle-infrastructure systems. Early datasets, such as KITTI \cite{Geiger2013-mu}, provided foundational support with a relatively simple sensor suite, including a frontal camera, LiDAR, and GNSS modules.Similarly, ApolloScape \cite{Huang2018}, Argoverse 2 \cite{wilson2023}, WHU-Urban3D \cite{HAN2024500}, ECMD \cite{Chen2024ECMD} and M2DGR \cite{YIN2022M2DGR} retained the core elements of LiDAR and GNSS while offering more extensive camera setups. 

Dataset focuses on GNSS/INS-only applications, such as the Google Smartphone Decimeter Challenge (SDC) \cite{google2022gsdc}, which provides raw GNSS measurements from smartphones across different scenarios, such as suburban areas and highways. A large number of data traces and testing area coverages provide a testbed for researchers to test their state-of-the-art algorithms to push the smartphone positioning accuracy to decimeter-level.

As the field evolved, later datasets began to incorporate more complex and varied sensor modalities. For instance, nuScenes\cite{Caesar2019-fh} and A2D2\cite{geyer2020} added 360-degree imaging and radar to better simulate real-world scenarios in both urban and suburban environments. These datasets improved sensor diversity and coverage.
The emergence of datasets like NTU4DRadLM \cite{zhang2023ntu4dradlm}  , 4D Radar Dataset \cite{Li2023radar}, and OmniHD \cite{zheng2025} marked a significant step toward more comprehensive sensing setups. These works introduced 4D radar sensors, capable of delivering velocity-embedded point clouds, and provided more varied terrain types, such as highways, rural, and campus environments. However, even these datasets still concentrate on the single vehicle preception and localization in urban environment. 

Several datasets have recently emerged with a focus on V2X applications, such as V2X-Sim \cite{Li9835036} and DAIR-V2X \cite{yu2022dairv2x}, which begin to bridge the gap by introducing infrastructure-based sensing concepts. The TUMTraf dataset\cite{zimmer2024tumtrafv2x} is one of the few that incorporates roadside LiDAR. However, it still lacks key sensing modalities such as panoramic vision and 4D radar, limiting its applicability for comprehensive cooperative perception and navigation research.

Our proposed dataset builds upon our released open-source platforms, UrbanLoco \cite{wen2020urbanloco} and UrbanNav \cite{Hsu2023UrbanNav}, and addresses their limitations by introducing a more comprehensive sensor suite. This includes both frontal and 360-degree cameras, LiDARs, 4D radar, GNSS raw measurements, and, critically, UWB modules deployed both onboard the vehicle and at roadside infrastructure. This configuration uniquely enables the study of infrastructure-assisted navigation and robust positioning in complex urban environments. The dataset has been preliminarily evaluated in our recent studies~\cite{HUANG2023RSILIO,HUANG2024errormap,HUANG2025RSGLIO}, demonstrating the effectiveness of cooperative navigation. Compared to prior efforts, our dataset is among the few that support full multisensory fusion while explicitly incorporating infrastructure-side sensing, providing a valuable benchmark for advancing V2X research and enabling real-world deployment in dense urban areas.


\begin{table*}[htbp]
        \renewcommand\arraystretch{1.2}
        \begin{center}
        \caption{Comparison of Sensor Setups for Existing and Proposed Datasets
        }
        \label{literature_comparison}
        \resizebox{\columnwidth*2}{!}
        { 
        \begin{threeparttable}
        \begin{tabular}{cccccccccccc} 
        \hline\hline  
        \multicolumn{1}{c}{\multirow{2}*{\textbf{Dataset}}}
    & \multicolumn{1}{c}{\multirow{2}*{\textbf{Terrain}}}
    & \multicolumn{2}{c}{\multirow{1}*{\textbf{Image}}}
    & \multicolumn{1}{c}{\multirow{2}*{\textbf{LiDAR}}}
    & \multicolumn{1}{c}{\multirow{2}*{\textbf{GNSS}}}
    & \multicolumn{1}{c}{\multirow{2}*{\textbf{Ground Truth}}}
    & \multicolumn{1}{c}{\multirow{2}*{\textbf{Radar}}}
    & \multicolumn{1}{c}{\multirow{2}*{\textbf{UWB}}}
    & \multicolumn{3}{c}{\multirow{1}*{\textbf{Roadside}}}
    \\
    \cline{3-4}
    \cline{10-12}
    & & \multicolumn{1}{c}{\multirow{1}*{Frontal}}
    & \multicolumn{1}{c}{\multirow{1}*{360-degree}}
    & & & & & & \multicolumn{1}{c}{\multirow{1}*{GNSS}}
    & \multicolumn{1}{c}{\multirow{1}*{LiDAR}}
    & \multicolumn{1}{c}{\multirow{1}*{UWB}}
    \\   
\hline
KITTI\cite{Geiger2013-mu}       & Urban     & \ding{51}     & \ding{55}     & \ding{51}    & \ding{51}  & \ding{51}    & \ding{55}      & \ding{55}     & \ding{55}     & \ding{55} & \ding{55} 
\\
nuScenes\cite{Caesar2019-fh}       & Urban, Suburban     & \ding{51}    & \ding{51}     & \ding{51}    & \ding{51}  & \ding{51}    & \ding{51}   & \ding{55}     & \ding{55}   & \ding{55}     & \ding{55}   \\
A2D2\cite{geyer2020}       & Urban, Rural, Off-Road     & \ding{51}    & \ding{51}     & \ding{51}    & \ding{51}  & \ding{55}    & \ding{55}   & \ding{55}     & \ding{55}         & \ding{55}     & \ding{55}\\

ApolloScape\cite{Huang2018}       & Urban     & \ding{51}    & \ding{51}     & \ding{51}    & \ding{51}   & \ding{51}     & \ding{55}  & \ding{55}    & \ding{55}   & \ding{55}     & \ding{55}    
\\
Argoverse 2\cite{wilson2023}       & Urban     & \ding{51}    & \ding{51}     & \ding{51}    & \ding{51}   & \ding{51}     & \ding{55}  & \ding{55}    & \ding{55}   & \ding{55}     & \ding{55} 
\\

WHU-Urban3D \cite{HAN2024500}       & Urban     & \ding{51}    & \ding{55}     & \ding{51}    & \ding{55}   & \ding{51}     & \ding{55}  & \ding{55}    & \ding{55}   & \ding{55}     & \ding{55} 
\\


4D Radar Dataset  \cite{Li2023radar}       & Campus, Industrial Zone     & \ding{51}    & \ding{55}     & \ding{51}    & \ding{51}   & \ding{51}     & 4D Radar  & \ding{55}    & \ding{55}   & \ding{55}     & \ding{55} 
\\

NTU4DRadLM \cite{zhang2023ntu4dradlm} & Campus     & \ding{51}    & \ding{55}     & \ding{51}    & \ding{51}   & \ding{51}     &   4D Radar   & \ding{55}    & \ding{55}      & \ding{55}  & \ding{55}    
\\
OmniHD\cite{zheng2025}       & Urban     & \ding{51}    & \ding{51}     & \ding{51}    & \ding{51}   & \ding{51}     & 4D Radar  & \ding{55}    & \ding{55}    & \ding{55}      & \ding{55}
\\

MCD \cite{nguyen2024mcd} & Campus     & \ding{51}    & \ding{55}     & \ding{51}    & \ding{51}   & \ding{51}     &  \ding{55}  & \ding{51}    & \ding{55}      & \ding{55}  & \ding{51}    
\\
UrbanLoco \cite{wen2020urbanloco}       & Urban & \ding{51}    &  \ding{51}     &  \ding{51}    &  NMEA+Raw    &  \ding{51}    &  \ding{55}    &  \ding{55}    &  \ding{55}    &  \ding{55}    &  \ding{55}
\\
UrbanNav \cite{Hsu2023UrbanNav}       & Urban     & \ding{51}    & \ding{55}     & \ding{51}    & NMEA+Raw   & \ding{51}     & \ding{55}  & \ding{55}    & \ding{55}    & \ding{55}    & \ding{55}
\\
V2X-Sim \cite{Li9835036}  & Simulated Urban, Suburban & \ding{51}    & \ding{51}     & \ding{51}    & \ding{55}   & \ding{51}     & \ding{55}  & \ding{55}    & \ding{55}    & \ding{51}    & \ding{55}
\\
DAIR-V2X \cite{yu2022dairv2x}        & Urban     & \ding{51}    & \ding{55}     & \ding{51}    & \ding{51}   & \ding{51}     & \ding{55}  & \ding{55}    & \ding{55}    & \ding{55}     & \ding{55}
\\
TUMTraf \cite{zimmer2024tumtrafv2x}        & Urban     & \ding{55}     & \ding{55}     & \ding{55}    & \ding{55}   & \ding{55}     & \ding{55}  & \ding{55}    & \ding{51} & \ding{51}   & \ding{55}     
\\
V2X-Real \cite{xiang2024v2xreal}  & Urban     & \ding{51}    & \ding{51}     & \ding{51}    & \ding{51}   & \ding{51}     & \ding{55}  & \ding{55}    & \ding{51}    & \ding{51}     & \ding{55}
 \\

UrbanV2X (proposed)     & Urban     & \ding{51}    & \ding{51}     & \ding{51}    &  NMEA+Raw  & \ding{51}      & 4D Radar  & \ding{51}    & \ding{51} & \ding{51} & \ding{51}    
\\
\hline\hline        
        \end{tabular}
        \end{threeparttable} 
        }
        \end{center}
\end{table*}

\section{System Overview}
\label{System Overview}


\subsection{Sensors Setup}
Our vehicle-Infrastructure platform is shown in Figure \ref{Fig_1_sensor_setup}. Onboard sensor-kit includes a multi-camera setup (two front-facing cameras with five surround cameras), two LiDARs, an onboard IMU, a vehicle-mounted UWB device, and a GNSS-RTK/INS system. Additionally, we have configured RSIs with its own GNSS, LiDAR, and UWB device synchronized with the vehicle.
The specific parameters of each sensor are listed in Table \ref{sensor_param}.

\subsubsection{Visual Sensors}
Seven Hikvision MV-CS050-10GC PRO industrial cameras equipped with VM0428MP12 4mm industrial lenses are used, each with a resolution of 2200$\times$1740, capturing RGB images at 10 Hz in fixed exposure mode. Among them, two are forward-facing stereo industrial cameras, while the other five cameras are installed at an angle of 60° between each other to form a 360° surround image capture setup. Each camera has a horizontal FOV of approximately 80.18°, with an overlap angle of around 20° between adjacent cameras. Image synchronization and triggering are achieved through the GigE Vision protocol, ensuring a frame trigger error of $< 0.005s$  for each image set.

\subsubsection{Mechanical LiDAR}
We have configured two mechanical LiDARs. A front-facing LiDAR, Hesai XT32, is tilted forward by 20° along the x-axis to collect precise road information ahead, including lane markings, traffic signs, and various road surface details, while also capturing architectural details of tall buildings. A central LiDAR, Velodyne HDL-32E, is mounted on the top of the vehicle to horizontally capture the surrounding environment. All LiDARs operate at a collection frequency of 10 Hz.

\subsubsection{4D Radar}
A 4D radar is mounted at the front of the sensor kit, capable of real-time measurement of the position, velocity, and distance of objects ahead, while generating 3D point cloud data fused with velocity information. It features a horizontal field of view of 150° and operates at a collection frequency of 13 Hz.

\subsubsection{Onboard UWB}
A Nooploop P-B UWB module is mounted onboard to provide short-range high-precision distance measurements with respect to roadside anchors and mobile platforms. It operates based on time-of-flight (ToF) measurements, offering centimeter-level ranging accuracy. UWB data are recorded at a frequency of 50 Hz, providing complementary positioning information to enhance system robustness in various environments.

\subsubsection{Onboard IMU and GNSS-RTK/INS suite}
An Xsens-MTI-30 IMU is mounted onboard to record raw acceleration and angular velocity data at a frequency of 400 Hz. Commercial-grade GNSS receivers with multi-constellation and dual-frequency support are included in the dataset. 

\subsubsection{Precise Reference System}
Precise localization ground truth is provided by a centimeter-level GNSS (supporting GPS and BeiDou) combined with an RTK/INS navigation system equipped with a fiber-optic gyroscope. Additional details are available in Section \ref{Ground-truth Poses}.

\subsubsection{RSIs}
 The RSIs are deployed along the testing route, equipped with a 300-line Innovusion Jaguar LiDAR operating at 10 Hz for high-resolution environmental perception. It is also equipped with an u-blox F9P GNSS receiver delivering roadside GNSS raw measurements and a Nooploop P-B UWB anchor to support precise relative localization. 

\begin{table}
\centering
\caption{Sensors specifications}
\label{sensor_param}
\resizebox{0.49\textwidth}{!}{
\begin{tabular}{cl} 
\hline\hline
Sensors                            & Specifications                                           \\ 
\hline
\multirow{4}{*}{Industrial Camera} & Hikvision MV-CS050-10GC PRO ($\times$ 7), 10 Hz          \\
                                   & 2200$\times$1740 pixel                                   \\
                                   & Lens: VM0428MP12 4mm                                     \\
                                   & H-FOV: 80.18$^\circ$, V-FOV: 73.77$^\circ$               \\ 
\hline
\multirow{8}{*}{LiDAR}             & Velodyne HDL-32E, 10Hz                                   \\
                                   & H-FOV: 360$^\circ$, V-FOV: +10.67$^\circ$$-30.67^\circ$  \\
                                   & 32 channel                                               \\
                                   & 100m range                                               \\ 
\cline{2-2}
                                   & Hesai XT32, 10Hz                                         \\
                                   & H-FOV: 360$^\circ$, V-FOV: +15$^\circ$$-16^\circ$        \\
                                   & 32 channel                                               \\
                                   & 120m range                                               \\ 
\hline
\multirow{3}{*}{4D Radar}          & GPAL-Ares-R7861, 13Hz                                    \\
                                   & Front-View                                               \\
                                   & H-FOV: 150$^\circ$, V-FOV: 30$^\circ$                    \\ 
\hline
\multirow{3}{*}{IMU}               & Xsens Mti-30, 400Hz                                      \\
                                   & Accelerometer in-run Bias Instability 15                 \\
                                   & Gyroscope in-run Bias Instability 18$^\circ$/h           \\ 
\hline
\multirow{2}{*}{GNSS Receiver}     & u-Blox ZED-F9P, 1Hz                                      \\
                                   & u-Blox EVK-F10T, 1Hz\\ 
\hline
\multirow{2}{*}{GNSS-RTK/INS}      & NovAtel SPAN-CPT , 1Hz                                   \\
                                   & Localization RMSE 5cm                                    \\ 
\hline
\multirow{6}{*}{Roadside Infrastruture}     & GNSS : u-Blox ZED-F9P, 1Hz                               \\
\cline{2-2}
                                   & UWB: Nooploop P-B, 50 Hz                                 \\ 
\cline{2-2}
                                   & LiDAR : Innovusion Jaguar LiDAR,10 Hz                    \\
                                   & H-FOV: 100$^\circ$, V-FOV: 40$^\circ$                                                  \\
                                   & 300 channel                                               \\
                                   & 280m range                                                         \\
\hline\hline
\end{tabular}
}
\end{table}

\subsection{System Architecture}

\begin{figure}[htb]  
        \centering
        \captionsetup{justification=justified}
        \includegraphics[width=0.8\columnwidth]{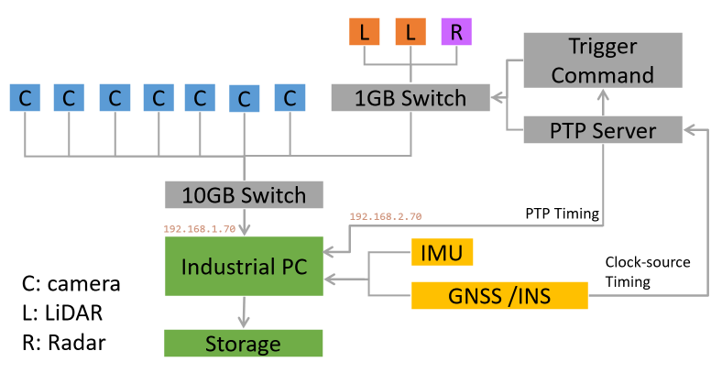}
        \caption{System Architecture}  
        \label{sys_architecture}
        \vspace{-2.0em}
\end{figure}

We use an Ethernet topology to facilitate data transmission, and the system architecture is illustrated in Fig. \ref{sys_architecture}. A Precision Time Protocol (PTP) \cite{PTP} server broadcasts synchronization signals to align the clocks of various data collection devices within the sensor network. The PTP server obtains NMEA \cite{langley1995nmea} output and pulse-per-second (PPS) signals from a u-blox M8T GNSS receiver, broadcasting synchronization data to all network ports. This aligns the ROS time of the onboard computer with GPS time and ensures hardware clock synchronization for cameras and LiDAR sensors. For sensors without Ethernet interfaces, timestamps are recorded using the GPS-synchronized ROS clock. The GigE Vision \cite{GigE} protocol is used to broadcast Action Commands to the network, achieving synchronized triggering of Ethernet cameras with a trigger accuracy of $< 0.005s$. All data is transmitted through this system architecture to onboard computers, where it is simultaneously recorded in rosbag format. Currently, the trigger frequency is set to 10Hz for LiDAR and Camera, 400Hz for IMU, 13Hz for 4D Radar, 1Hz for GNSS, and 50Hz for UWB.

\subsection{Sensors Calibration}

\begin{figure}[htb]  
        \centering
        \captionsetup{justification=justified}
        \includegraphics[width=0.8\columnwidth]{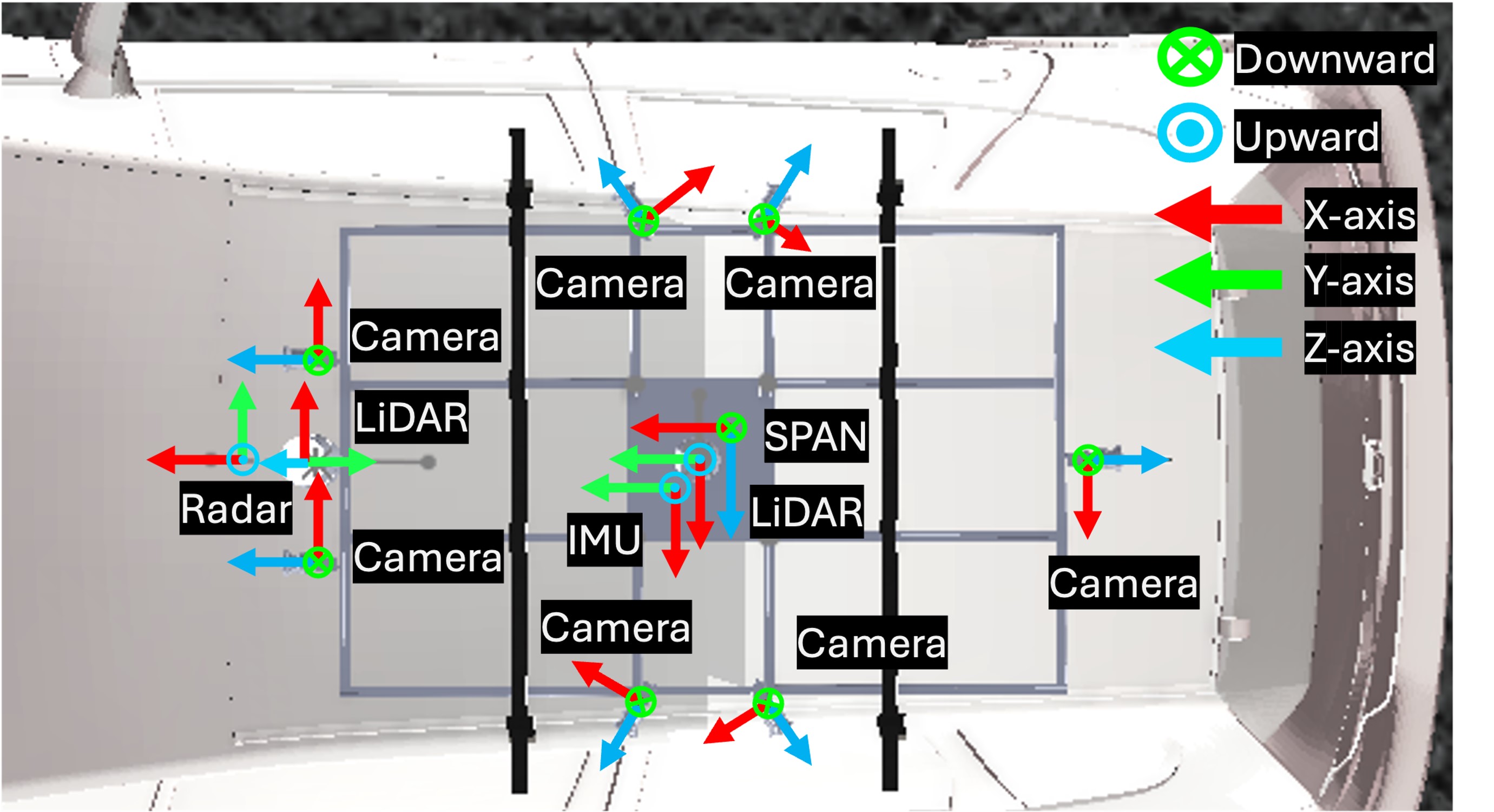}
        \caption{Sensor Frame}  
        \label{ex_frame}
        \vspace{-0.0em}
\end{figure}

The structure of all sensors is quantitatively designed, with their coordinate frames shown in Fig. \ref{ex_frame}, and the extrinsic parameters of the sensors are calibrated in practice.

\subsubsection{LiDAR-Camera Calibration}
\begin{figure}[htb]  
    	\subfigtopskip=0pt 
    	\subfigbottomskip=0pt 
    	\subfigcapskip=-8pt 
        \captionsetup{justification=justified}
        \centering
        \subfigure[]{
                \begin{minipage}[t]{0.48\columnwidth}
                \centering
                \includegraphics[width=1.0\columnwidth]{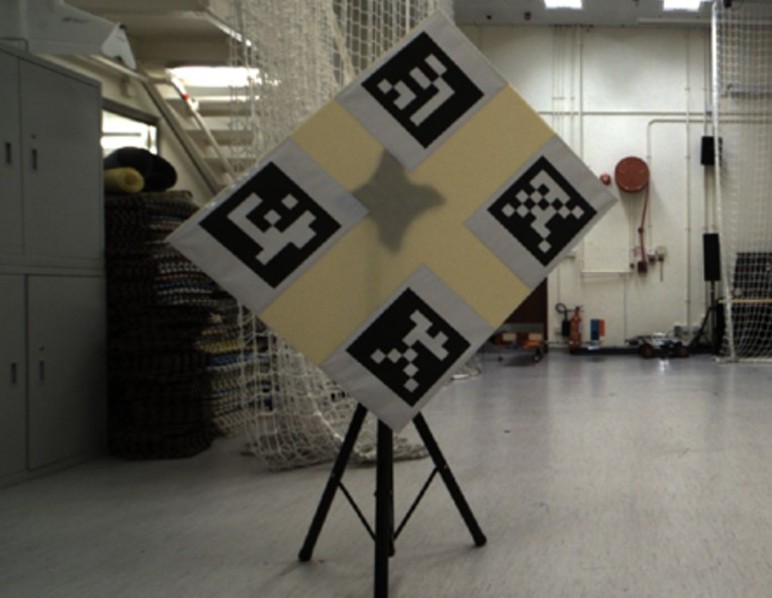}
                \label{lc_cal_board}
                \end{minipage}%
        }%
        \subfigure[]{
                \begin{minipage}[t]{0.48\columnwidth}
                \centering
                \includegraphics[width=1.0\columnwidth]{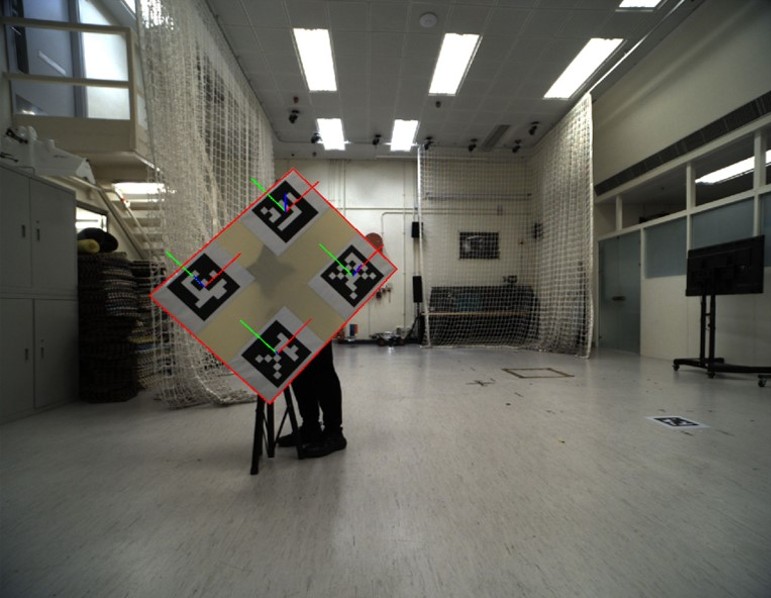}
                \label{lc_cal_2}
                \end{minipage}%
        } \\ %
        \subfigure[]{
                \begin{minipage}[t]{0.48\columnwidth}
                \centering
                \includegraphics[width=1.0\columnwidth]{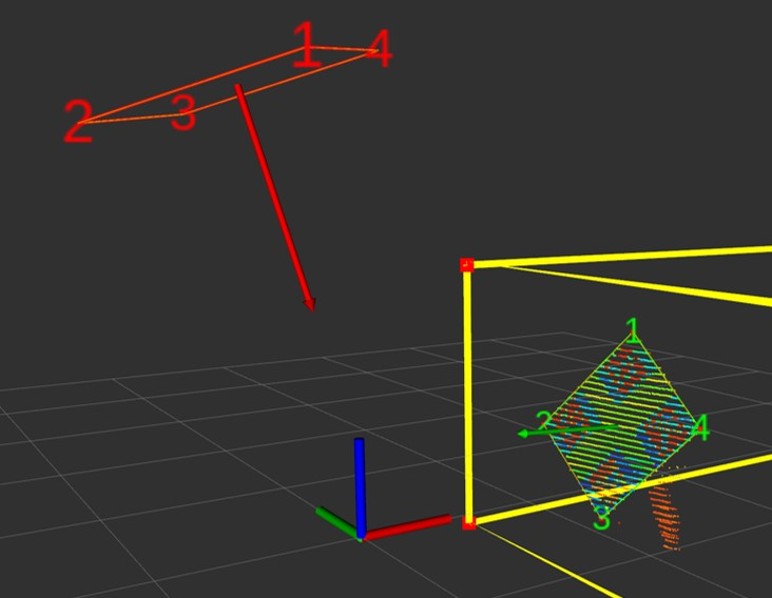}
                \label{lc_cal_1}
                \end{minipage}%
        }%
        \subfigure[]{
                \begin{minipage}[t]{0.48\columnwidth}
                \centering
                \includegraphics[width=1.0\columnwidth]{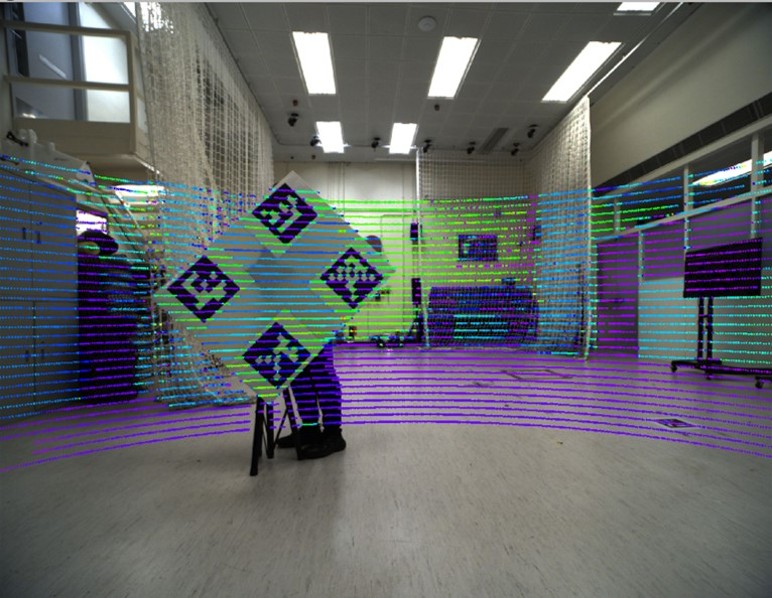}
                \label{lc_cal_3}
                \end{minipage}%
        }
        \caption{The calibration between LiDAR and Camera. (a) Custom-Designed ArUco Calibration Board. ArUco markers are symmetrically attached around the rectangular calibration board. (b) The corners extracted from the image and point cloud in 3D space. (c) Extracting the 3D corner points of the calibration board from the image. The green parts represent the segmented calibration board point cloud plane and corner points, while the red parts represent the calibration board corner points calculated from the ArUco markers detected in the image.  (d) Image fused with point cloud. The fusion effect is verified by projecting the point cloud onto the image.}
        \label{lc_cal}
\end{figure}

We use an Iterative Closest Point (ICP)-based method \cite{icp1992} with a custom-designed ArUco calibration board for calibration, as shown in Fig. \ref{lc_cal_board}. In the image data, the positions and poses of multiple ArUco markers are detected, and quaternion averaging is applied to estimate their positions and poses, ultimately determining the 3D pose of the calibration board and the locations of its corner points in 3D space, as illustrated in Fig. \ref{lc_cal_2}. In the point cloud data, the calibration board's cluster is extracted using normal-based clustering, and plane parameters are estimated with random sample consensus (RANSAC). The corner points of the calibration board are then extracted in the 2D plane using the minimum bounding rectangle method. These corner points are subsequently mapped back to the original 3D plane based on the plane parameters, resulting in the 3D corner points of the calibration board. The extraction of corner points from both the image and the point cloud is depicted in Fig. \ref{lc_cal_1}. Finally, the LiDAR-Camera extrinsic matrix is optimized and estimated through ICP. The point cloud can be then well projected onto the image using the extrinsic matrix and the camera's intrinsic parameters, as shown in Fig. \ref{lc_cal_3}. By performing calibration in this manner and analyzing the reprojection error, the resulting calibration error is approximately 2.3 pixels.

\subsubsection{Camera-IMU Calibration}
We used checkerboards as calibration references and moved the sensor kit in front of the checkerboards to ensure the cameras captured the patterns while providing sufficient excitation for the IMU, simultaneously collecting data. Subsequently, Kalibr \cite{furgale2013unified} was utilized to estimate the extrinsics and time offsets between all cameras and the IMU.

\subsubsection{LiDAR-IMU Calibration}
For mechanical LiDAR calibration, LI-Init \cite{LiDARCalibrMars2022} can perform spatiotemporal calibration between the LiDAR and IMU without requiring checkerboards or additional equipment, as illustrated in Fig. \ref{li_cal}. The device is moved and rotated along the XYZ axes to generate adequate excitation. Once sufficient data has been collected, the extrinsic parameters between the LiDAR and IMU are derived.


\subsubsection{LiDAR-LiDAR Calibration}
\begin{figure}[htb]  
    	\subfigtopskip=0pt 
    	\subfigbottomskip=0pt 
    	\subfigcapskip=-8pt 
        \captionsetup{justification=justified}
        \centering
        \subfigure[]{
                \begin{minipage}[t]{0.48\columnwidth}
                \centering
                \includegraphics[width=1.0\columnwidth]{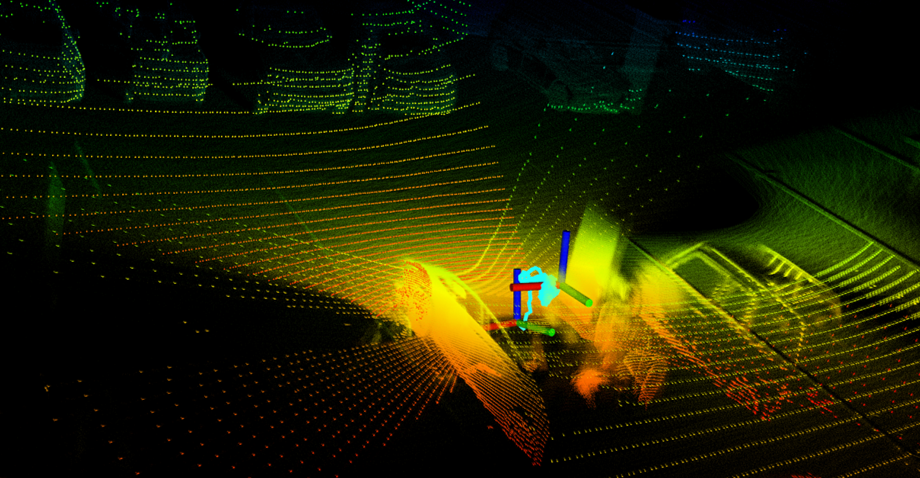}
                \label{li_cal}
                \end{minipage}%
        }%
        \subfigure[]{
                \begin{minipage}[t]{0.48\columnwidth}
                \centering
                \includegraphics[width=1.0\columnwidth]{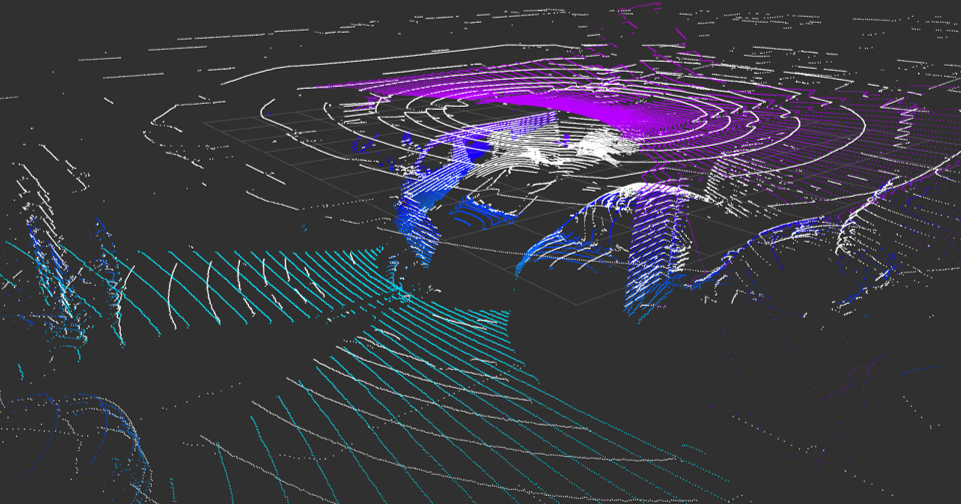}
                \label{ll_cal}
                \end{minipage}%
        } \\ %
        \caption{The extrinsics calibration for LiDAR and IMU. (a) LiDAR-IMU calibration. The image illustrates the process of sensor motion calibration using the LI-Init \cite{LiDARCalibrMars2022} method. The point cloud shows the constructed point cloud structure, while the blue line represents the motion trajectory. (b) LiDAR-LiDAR calibration. The image shows the fused point cloud after the registration of two LiDARs, where the white point cloud represents the Velodyne data, and the colored point cloud represents the Hesai data.}
        \label{lc_cal}
\end{figure}

The calibration of LiDAR is performed using the Normal Distributions Transform (NDT) registration method \cite{Martin2009}. First, VoxelGrid filtering is applied to voxelize the point cloud, followed by Statistical Outlier Removal (SOR) to eliminate outliers. Using the extrinsics obtained from LiDAR-IMU calibration, the relative extrinsic parameters between the two LiDARs are derived through inverse transformation as the initial guess. Based on this, NDT is used to register the two sets of point clouds and obtain the extrinsic matrix, shown as Fig. \ref{ll_cal}.

\subsubsection{LiDAR/Camera-4DRadar Calibration}
To perform extrinsic calibration between the LiDAR, camera, and 4D radar, we referenced the method proposed in \cite{Radar2ThermalCalib}. Specifically, a calibration board embedded with a corner reflector was created, with the center of the corner reflector serving as the radar feature. The LiDAR uses the method mentioned in 1) to extract the center point of the board. For the image, the center point of the board is extracted using ArUco markers. Finally, the extrinsic calibration between LiDAR-Radar and Camera-Radar is achieved using the ICP method.

\section{Dataset Overview}
\label{Dataset Overview}

\begin{figure}[htb]  
        \centering
        \captionsetup{justification=justified}
        \includegraphics[width=0.9\columnwidth]{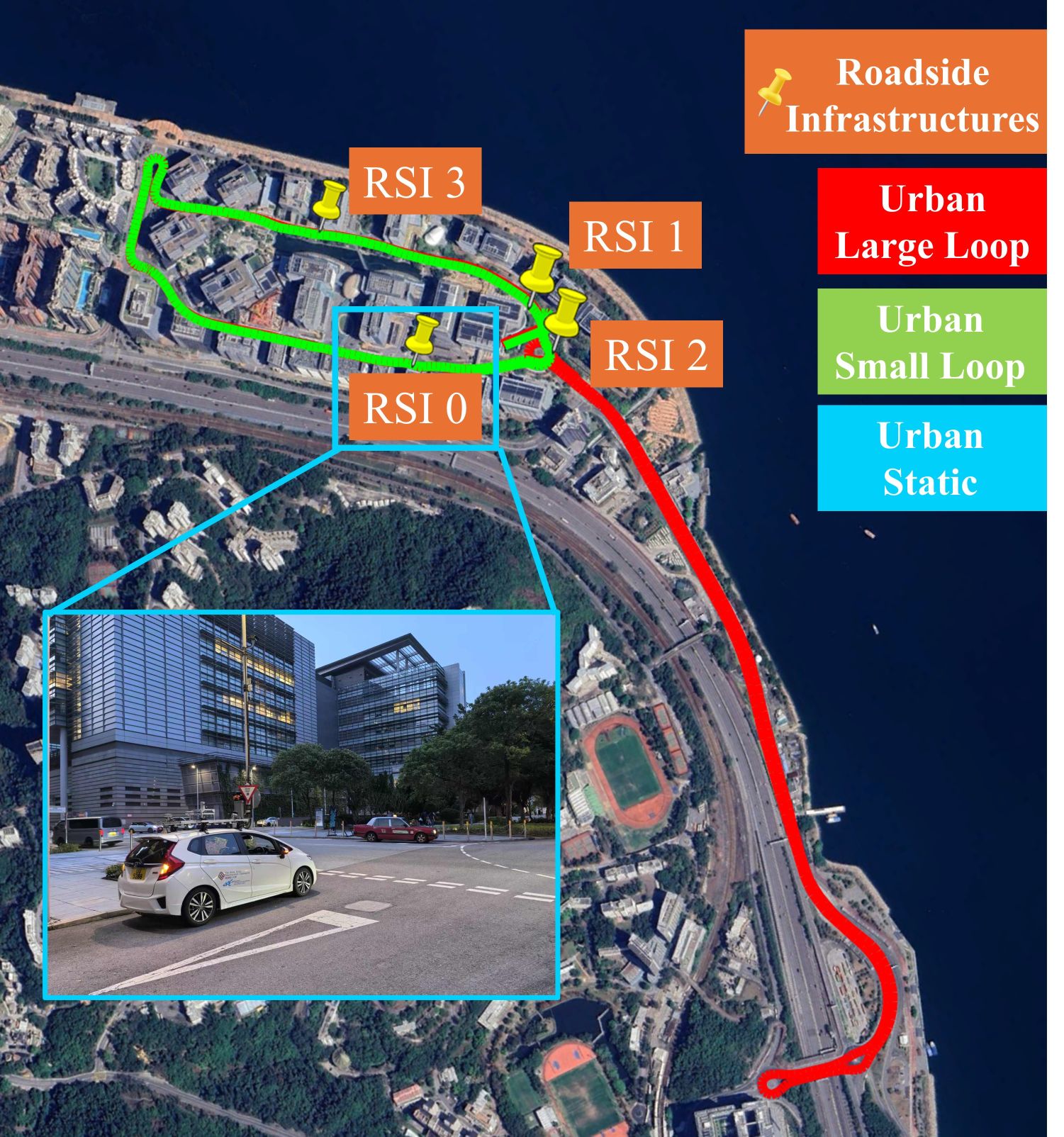}
        \caption{Illustration of three scenarios and RSI platform, including both loop and static scenarios.}  
        \label{route}
        \vspace{-1.0em}
\end{figure}

As shown in Fig. \ref{route}, our dataset covers a wide range of driving scenarios in the Hong Kong C-V2X testbed, including urban streets, city roads, and highways.

\begin{figure*}[htb]  
        \centering
        \captionsetup{justification=justified}
        \includegraphics[width=2.0\columnwidth]{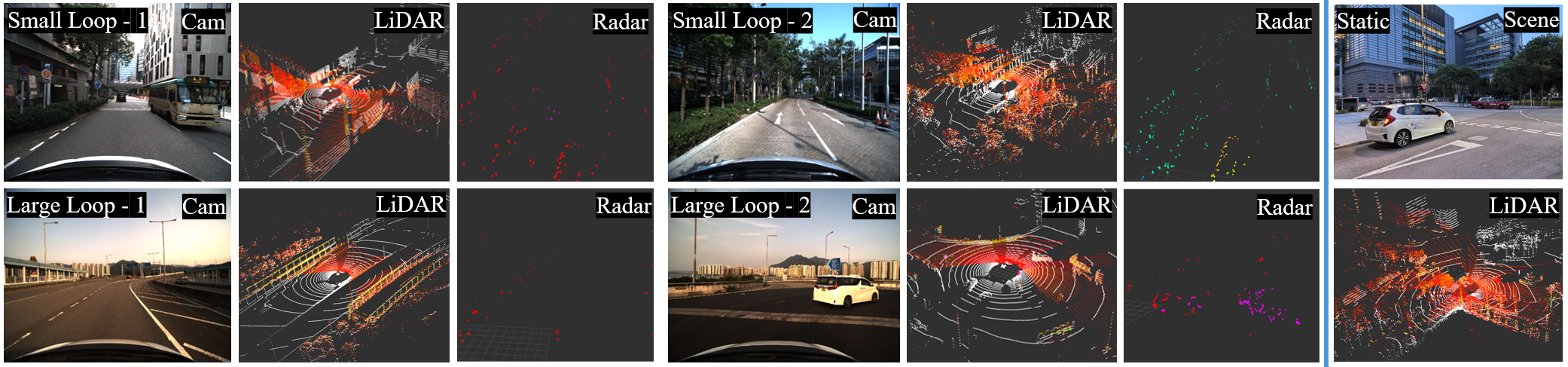}
        \caption{Visualization of the three scenarios: the first row shows a typical scene from the \textit{Urban-Small-Loop}, the second row represents the \textit{Urban-Large-Loop}, and the part to the right of the blue vertical line represents the \textit{Urban-Static} scenario. }  
        \label{Scene}
        \vspace{-1.0em}
\end{figure*}

As shown in the \ref{table_summary_of_dataset}, our dataset includes Small Loop, Large Loop, and Static scenarios, with preview images of the data scenes shown in Fig. \ref{Scene}. The Small Loop primarily includes dense urban roads, open areas, and narrow areas with high vehicle density. The time distribution is limited to daytime hours under clear weather conditions. The Large Loop, in addition to covering the scenarios in the Small Loop, also includes highway segments and busy road sections. The Static data consists of 30 minutes of stationary data collected while parked beside a roundabout in the scene, with long-term coverage of roadside UWB data nearby.


\begin{table}[htbp]
    \setlength{\abovecaptionskip}{-0.02cm}
    \renewcommand\arraystretch{1.2}
    \tiny 
    \begin{center}
    \caption{Summary of Dataset}
    \label{table_summary_of_dataset}
    \resizebox{\columnwidth*1}{!}
    {
    \begin{tabular}{p{1cm} p{1cm}  p{1cm} p{1cm}}
        \hline
        \textbf{Scenarios} & \textbf{Terrain} & \textbf{Duration (s)} & \textbf{Size (GB)} \\ \hline
        \multirow{1}{*}{Small-Loop} 
        & Urban Road   & 309 & 27\\
         \hline
      
        \multirow{1}{*}{Large-Loop} 
        & Urban Road, Highway   & 600 & 48.6 \\ \hline

        \multirow{1}{*}{Static} 
        & /             & 1800 & / \\ \hline
    \end{tabular}
    }
    \end{center}
\end{table}


\subsection{Scenarios}

\subsubsection{Urban-Small-Loop}

The small-loop scenario is conducted within the C-V2X testbed, with each loop spanning approximately 1.8 kilometers. This area represents a densely built urban canyon environment, featuring narrow streets flanked by closely spaced high-rise buildings. The vehicle operates at a relatively low speed of around 35 km/h to mimic typical urban traffic conditions. To enable cooperative localization, multiple RSIs are deployed along the route. Each RSI is equipped with a LiDAR, a UWB anchor, a GNSS receiver, and a synchronized data logger to capture UWB ranging measurements and satellite data in real-time. These RSIs are positioned at elevated locations to minimize signal blockage and enhance line-of-sight coverage. The scenario is specifically designed to evaluate loop closure performance in SLAM systems by driving the vehicle along a repetitive path over multiple loops.

\subsubsection{Urban-Large-Loop}

The large-loop scenario spans from HKSTP to a nearby university campus, covering approximately 5.8 kilometers per loop. This route includes both complex urban canyon segments and open highway sections, providing a diverse range of localization challenges. In the urban segments, the vehicle travels through areas with high-rise buildings and dense traffic at speeds under 40 km/h, while in the highway sections, it accelerates to around 60 km/h along broader, multi-lane roads. The reduced feature richness and faster dynamics in the highway portions pose additional challenges for vision- and LiDAR-based localization systems. Similar to the small-loop setting, RSIs are strategically installed along the route. These RSIs share a common configuration, consisting of LiDARs, UWB anchors and GNSS-RTK modules synchronized to a common time base, enabling high-accuracy, low-latency cooperative localization support across the entire loop.
\subsubsection{Urban-Static}
In the static scenario, the vehicle is parked near a roadside GNSS reference station for 30 minutes without any movement. This setup aims to assess the contribution of infrastructure-assisted positioning, especially in mitigating multipath errors and improving GNSS accuracy under short baseline conditions. The static deployment benefits from close proximity to a fixed GNSS-RTK base station, enabling centimeter-level correction through direct reference positioning, which is essential for evaluating the baseline drift and positioning consistency over extended durations.

%

\subsubsection{Ground-truth Poses}
\label{Ground-truth Poses}
To obtain high-accuracy ground-truth trajectories, we employ a tightly coupled GNSS-RTK/INS system \cite{NovAtel} that fuses multi-constellation GNSS measurements with a fiber-optic gyroscope-based inertial navigation system. This setup provides centimeter-level positioning accuracy even in challenging urban environments. The system outputs 6-DoF pose estimates (position and orientation) at 1 Hz, serving as reliable reference trajectories for evaluating SLAM and localization algorithms. 

\subsubsection{LiDAR Color Point Cloud Maps Generation}
\label{LiDAR Color Point Cloud Maps Generation}

\begin{figure}[htb]  
        \centering
        \captionsetup{justification=justified}
        \includegraphics[width=0.9\columnwidth]{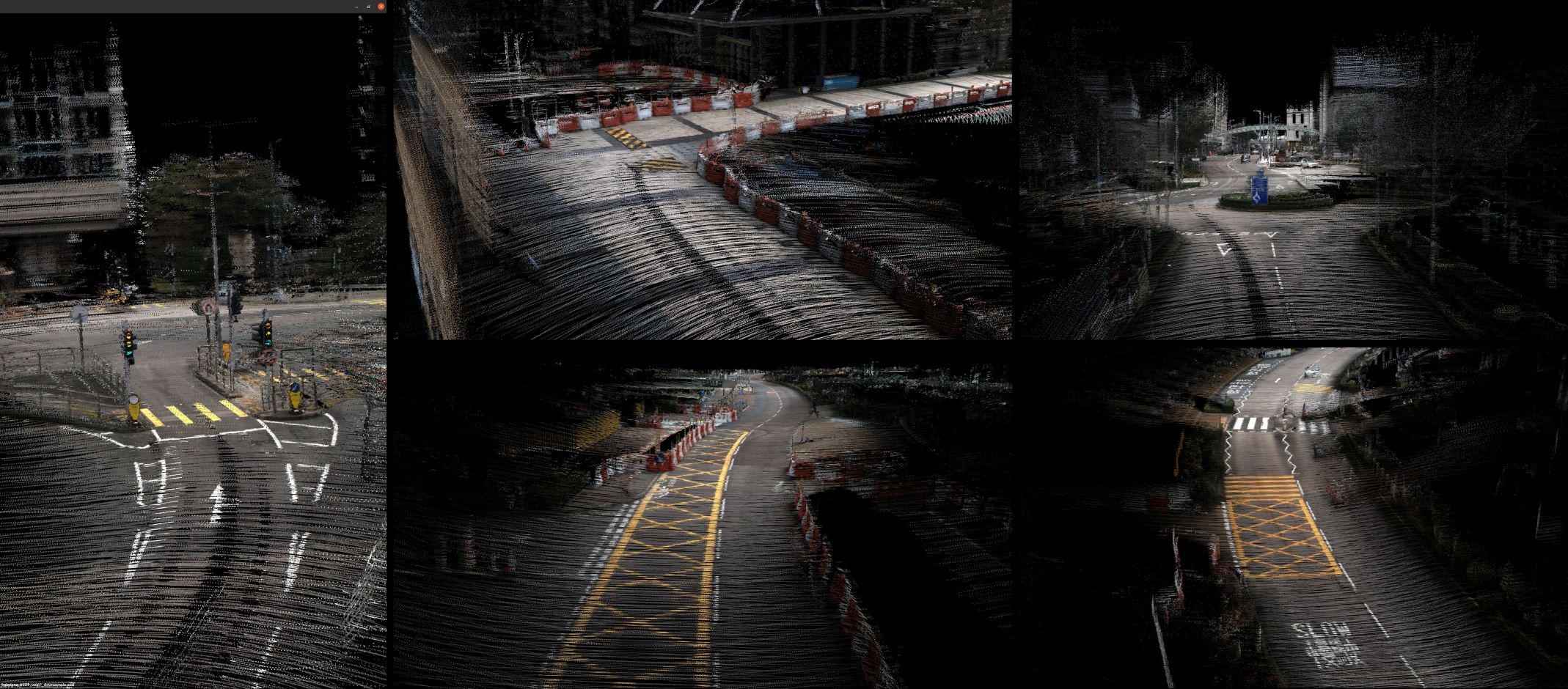}
        \caption{The color point cloud in Urban-Small-Loop scene.}  
        \label{color_cloud}
        \vspace{-1.0em}
\end{figure}

The Fast-LIVO2 \cite{FAST-LIVO2} method is used to construct colored point clouds of road scenes. By configuring the extrinsic and time synchronization of the LiDAR, IMU, and Camera, colored point cloud SLAM can be performed directly, resulting in the colored point cloud scene shown in Fig. \ref{color_cloud}, which features clear and rich color detail characteristics.

\section{Dataset Evaluation}
\label{Dataset Evaluation}

\subsection{Onboard Evaluations}
\label{Visual SLAM Evaluations}
\begin{table}[htbp]
    \setlength{\abovecaptionskip}{-0.02cm}
    \renewcommand\arraystretch{1.2}
    \tiny 
    \begin{center}
    \caption{Performance evaluation of listed methods using onboard platform.}
    \label{table_evaluation_onboard}
    \resizebox{\columnwidth*1}{!}
    {
    \begin{tabular}{p{1cm} p{1cm}  p{1cm}  p{1cm}}
        \hline
        \textbf{Scenarios} & \textbf{Methods} & \textbf{APE RMSE (m)} & \textbf{APE MEAN (m)} \\ \hline
        \multirow{5}{*}{Urban-Small-Loop} 
        & VINS-Mono Front-left-2   & 61.568 & 51.595 \\
        & VINS-Mono Front-right-1   & 166.178 & 143.319 \\
        & LIO-SAM           & 18.851 & 16.277 \\
        & Fast-LIO2          & 29.311 & 25.693 \\
        & Fast-LIVO2           & 304.246 & 260.716  \\
         & GNSS-RTK          &18.595 &  10.973\\ \hline
      
        \multirow{5}{*}{Urban-Large-Loop} 
        & VINS-Mono Front-left-2   & 184.866 & 165.212 \\
        & VINS-Mono Front-right-1   & Failed & Failed \\
        & LIO-SAM           & 49.985 & 57.465 \\
        & Fast-LIO2          & 43.284 & 32.904 \\
        & Fast-LIVO2           & 379.946 & 320.118  \\
         & GNSS-RTK          & 12.443 & 6.764\\ \hline

        \multirow{1}{*}{Urban-Static} 
        &GNSS-RTK             & 14.986 & 13.999  \\ \hline
    \end{tabular}
    }
    \end{center}
\end{table}

As shown in Table \ref{table_evaluation_onboard}, we evaluated the performance of visual SLAM and LiDAR SLAM without loop closure in both Large Loop and Small Loop scenarios. For \textbf{visual SLAM}, we used VINS-MONO \cite{VINS-MONO} with forward-facing left and right cameras. It performed well in the \textbf{Small Loop} scenario due to high-precision time synchronization and high-quality images but failed in the \textbf{Large Loop} scenario, primarily due to motion blur, feature sparsity, cumulative drift, and errors caused by vehicle vibrations and high-speed motion. For \textbf{LiDAR SLAM}, Fast-LIO2 \cite{Fast-lio2} and LIO-SAM \cite{Lio-sam} showed better accuracy in the Small Loop and Large Loop than Fast-LIVO2 \cite{FAST-LIVO2}, particularly in feature-sparse highway scenarios that are prone to long-distance drift and susceptible to errors in height estimation. \textbf{GNSS-RTK}, evaluated using RTKLIB \cite{takasu2009RTKLIB} with correction data from the Hong Kong Observatory base station, achieved the better accuracy across all scenarios. These results highlight the challenging nature of the \textbf{UrbanV2X dataset}, making it a valuable benchmark for advancing positioning and SLAM research.


\subsection{Roadside GNSS Evaluations}
Table \ref{evaluation_roadside_GNSS} presents the positioning accuracy results for the \textit{Urban-Static} scenario. Using RTKLIB \cite{takasu2009RTKLIB} with the traditional configuration—relying on a publicly available Hong Kong Observatory base station, the RMSE is 14.986 m. In contrast, GNSS-RTK-
RSG, which uses a nearby roadside GNSS unit as the differential reference, achieves a significantly lower RMSE of 8.422 m. This substantial improvement highlights the benefits of incorporating roadside GNSS, which offers a shorter baseline and experiences similar multipath conditions \cite{Zhang2020collaboration}.

\begin{table}[htbp]
    \setlength{\abovecaptionskip}{-0.02cm}
    \renewcommand\arraystretch{1.2}
    \tiny 
    \begin{center}
    \caption{Performance evaluation of GNSS positioning methods using tradition station (GNSS-RTK) and roadside GNSS (GNSS-RTK-RSG)}
    \label{evaluation_roadside_GNSS}
    \resizebox{\columnwidth*1}{!}
    {
    \begin{tabular}{p{1cm} p{1cm}  p{1cm}  p{1cm}}
        \hline
        \textbf{Sequence} & \textbf{Methods} & \textbf{APE RMSE (m)} & \textbf{APE MEAN (m)} \\ \hline

        \multirow{1}{*}{Urban-Static} 
          & GNSS-RTK           & 14.986 & 13.999 \\
 
         & GNSS-RTK-RSG     &8.422& 8.152   \\ \hline
    \end{tabular}
    }
    \end{center}
     \vspace{-1.0em}

\end{table}

\subsection{Roadside UWB Evaluations}
\begin{figure}[htb]  
        \centering
        \captionsetup{justification=justified}
        \includegraphics[width=1.0\columnwidth]{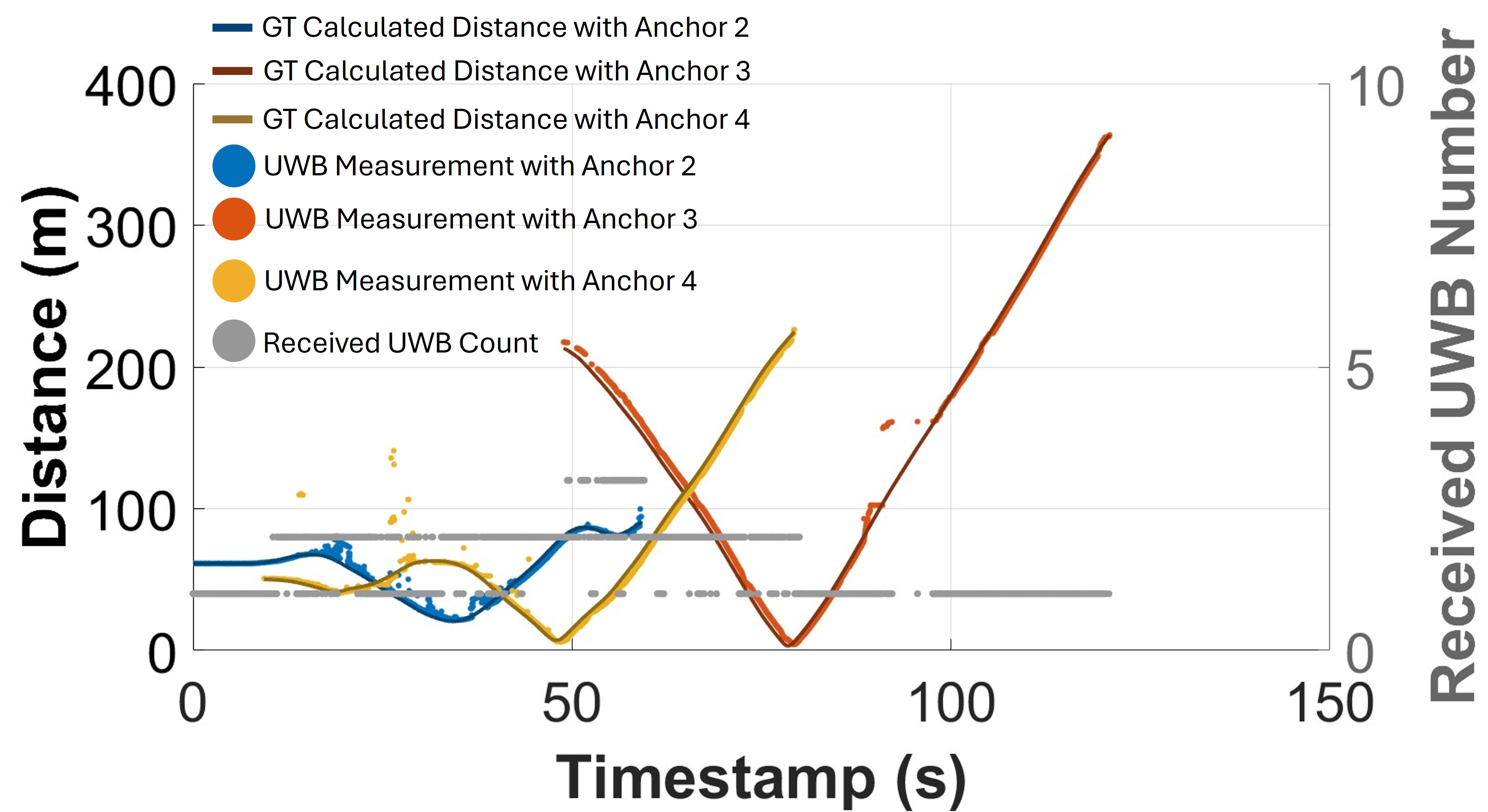}
        \caption{The relative range between UWB measured distance with ground truth, and the number of received UWB in small loop}  
        \label{uwb_residual}
        \vspace{-0.0em}
\end{figure}

As shown in Fig. \ref{route}, we deployed 3 UWB anchors with RSI, which ensure the communication between vehicle and RSU at the roundabout area.   The experimental results, as shown in Fig. \ref{uwb_residual}, shows that the RSIs with UWB can provide high localization accuracy within detection range, which can serve as a robust supplement to conventional localization technologies. This makes it a viable option for supporting cooperative localization in V2X applications, especially where GNSS reliability is compromised.

\section{Conclusions} 
\label{CONCLUSIONS}
We presented UrbanV2X, a comprehensive multisensory dataset tailored for vehicle–infrastructure cooperative navigation in dense urban environments. The dataset includes time-synchronized and calibrated data from both onboard and roadside sensors—LiDAR, 4D radar, UWB, industrial cameras, and GNSS-RTK/INS—enabling research on cross-view perception and infrastructure-assisted localization. Preliminary evaluations using state-of-the-art SLAM and positioning algorithms demonstrate the advantages of incorporating infrastructure sensing for enhanced robustness and accuracy.

UrbanV2X provides a unique testbed for developing and benchmarking cooperative autonomy solutions under real-world constraints such as dynamic objects, GNSS degradation, and occlusions. We believe this dataset will support future research in V2X perception, multi-agent navigation, and smart city applications. The dataset is publicly available to promote reproducible research and accelerate progress in intelligent transportation systems.

In future work, we will further enhance the richness of the data and the completeness of the evaluations. We will also adopt various collaborative SLAM methods \cite{DCL-SLAM2024,Schmuck2019,Schmuck2017} for validation. Additionally, we will continue working on annotations to make the work more valuable.

\section*{Acknowledgments}
The authors thank Yuteng Wang, Naigui Xiao, Shaoting Qiu from PolyU, Stella Zhu, Jiashi Feng, Alpamys Urtay and Siqiao Wu from ASTRI for their kind support in this data evaluation. 





\bibliographystyle{IEEEtran} 
\bibliography{references.bib} 

@string{iros = "{IEEE/RSJ Int. Conf. on Robots and Intelligent Systems}"}

@string{icra = "{IEEE Int. Conf. on Robotics and Automation}"}

@ARTICLE{Chen2024ECMD,
  author={Chen, Peiyu and Guan, Weipeng and Huang, Feng and Zhong, Yihan and Wen, Weisong and Hsu, Li-Ta and Lu, Peng},
  journal={IEEE Transactions on Intelligent Vehicles}, 
  title={ECMD: An Event-Centric Multisensory Driving Dataset for SLAM}, 
  year={2024},
  volume={9},
  number={1},
  pages={407-416},
  doi={10.1109/TIV.2023.3339144}}

@ARTICLE{Zhang2020collaboration,
  author={Zhang, Guohao and Ng, Hoi-Fung and Wen, Weisong and Hsu, Li-Ta},
  journal={IEEE Transactions on Intelligent Transportation Systems}, 
  title={3D Mapping Database Aided GNSS Based Collaborative Positioning Using Factor Graph Optimization}, 
  year={2021},
  volume={22},
  number={10},
  pages={6175-6187},
  doi={10.1109/TITS.2020.2988531}}

@inproceedings{takasu2009RTKLIB,
  title={Development of the low-cost RTK-GPS receiver with an open source program package RTKLIB},
  author={Takasu, Tomoji and Yasuda, Akio},
  booktitle={International symposium on GPS/GNSS},
  volume={1},
  pages={1--6},
  year={2009},
  organization={International Convention Center Jeju Korea Seogwipo-si, Republic of Korea}
}

@inproceedings{Hsu2023UrbanNav,
  title={Hong Kong UrbanNav: An open-source multisensory dataset for benchmarking urban navigation algorithms},
  author={Hsu, Li-Ta and Huang, Feng and Ng, Hoi-Fung and Zhang, Guohao and Zhong, Yihan and Bai, Xiwei and Wen, Weisong},
  journal={NAVIGATION: Journal of the Institute of Navigation},
  volume={70},
  number={4},
  year={2023},
  publisher={Institute of Navigation}
}

@ARTICLE{Geiger2013-mu,
  title    = "Vision meets robotics: The {KITTI} dataset",
  author   = "Geiger, Andreas and Lenz, Philip and Stiller, C and Urtasun, R",
  journal  = "The International Journal of Robotics Research",
  volume   =  32,
  pages    = "1231--1237",
  year     =  2013
}

@ARTICLE{Caesar2019-fh,
  title    = "nuScenes: A Multimodal Dataset for Autonomous Driving",
  author   = "Caesar, Holger and Bankiti, Varun and Lang, Alex H and Vora,
              Sourabh and Liong, Venice Erin and Xu, Qiang and Krishnan, Anush
              and Pan, Yuxin and Baldan, G and Beijbom, Oscar",
  journal  = "Proc. IEEE Comput. Soc. Conf. Comput. Vis. Pattern Recognit.",
  pages    = "11618--11628",
  year     =  2019
}

@misc{geyer2020,
      title={A2D2: Audi Autonomous Driving Dataset}, 
      author={Jakob Geyer and Yohannes Kassahun and Mentar Mahmudi and Xavier Ricou and Rupesh Durgesh and Andrew S. Chung and Lorenz Hauswald and Viet Hoang Pham and Maximilian Mühlegg and Sebastian Dorn and Tiffany Fernandez and Martin Jänicke and Sudesh Mirashi and Chiragkumar Savani and Martin Sturm and Oleksandr Vorobiov and Martin Oelker and Sebastian Garreis and Peter Schuberth},
      year={2020},
      eprint={2004.06320},
      archivePrefix={arXiv},
      primaryClass={cs.CV},
      url={https://arxiv.org/abs/2004.06320}, 
}

@misc{zheng2025,
      title={OmniHD-Scenes: A Next-Generation Multimodal Dataset for Autonomous Driving}, 
      author={Lianqing Zheng and Long Yang and Qunshu Lin and Wenjin Ai and Minghao Liu and Shouyi Lu and Jianan Liu and Hongze Ren and Jingyue Mo and Xiaokai Bai and Jie Bai and Zhixiong Ma and Xichan Zhu},
      year={2025},
      eprint={2412.10734},
      archivePrefix={arXiv},
      primaryClass={cs.CV},
      url={https://arxiv.org/abs/2412.10734}, 
}

@INPROCEEDINGS{Huang2018,
  author={Huang, Xinyu and Cheng, Xinjing and Geng, Qichuan and Cao, Binbin and Zhou, Dingfu and Wang, Peng and Lin, Yuanqing and Yang, Ruigang},
  booktitle={2018 IEEE/CVF Conference on Computer Vision and Pattern Recognition Workshops (CVPRW)}, 
  title={The ApolloScape Dataset for Autonomous Driving}, 
  year={2018},
  volume={},
  number={},
  pages={1067-10676},
  doi={10.1109/CVPRW.2018.00141}
}

@misc{wilson2023,
      title={Argoverse 2: Next Generation Datasets for Self-Driving Perception and Forecasting}, 
      author={Benjamin Wilson and William Qi and Tanmay Agarwal and John Lambert and Jagjeet Singh and Siddhesh Khandelwal and Bowen Pan and Ratnesh Kumar and Andrew Hartnett and Jhony Kaesemodel Pontes and Deva Ramanan and Peter Carr and James Hays},
      year={2023},
      eprint={2301.00493},
      archivePrefix={arXiv},
      primaryClass={cs.CV},
      url={https://arxiv.org/abs/2301.00493}, 
}

@article{HAN2024500,
title = {WHU-Urban3D: An urban scene LiDAR point cloud dataset for semantic instance segmentation},
journal = {ISPRS Journal of Photogrammetry and Remote Sensing},
volume = {209},
pages = {500-513},
year = {2024},
issn = {0924-2716},
doi = {https://doi.org/10.1016/j.isprsjprs.2024.02.007},
url = {https://www.sciencedirect.com/science/article/pii/S0924271624000522},
author = {Xu Han and Chong Liu and Yuzhou Zhou and Kai Tan and Zhen Dong and Bisheng Yang},
}

@ARTICLE{Li2023radar,
  author={Li, Xingyi and Zhang, Han and Chen, Weidong},
  journal={IEEE Robotics and Automation Letters}, 
  title={4D Radar-Based Pose Graph SLAM With Ego-Velocity Pre-Integration Factor}, 
  year={2023},
  volume={8},
  number={8},
  pages={5124-5131},
  doi={10.1109/LRA.2023.3292574}
}

@INPROCEEDINGS{wen2020urbanloco,
  author={Wen, Weisong and Zhou, Yiyang and Zhang, Guohao and Fahandezh-Saadi, Saman and Bai, Xiwei and Zhan, Wei and Tomizuka, Masayoshi and Hsu, Li-Ta},
  booktitle={2020 IEEE International Conference on Robotics and Automation (ICRA)}, 
  title={UrbanLoco: A Full Sensor Suite Dataset for Mapping and Localization in Urban Scenes}, 
  year={2020},
  volume={},
  number={},
  pages={2310-2316},
  doi={10.1109/ICRA40945.2020.9196526}
}

@ARTICLE{Bai2021DegeneratedVIO,
  author={Bai, Xiwei and Wen, Weisong and Hsu, Li-Ta},
  journal={IEEE Transactions on Instrumentation and Measurement}, 
  title={Degeneration-Aware Outlier Mitigation for Visual Inertial Integrated Navigation System in Urban Canyons}, 
  year={2021},
  volume={70},
  number={},
  pages={1-15},
  doi={10.1109/TIM.2021.3126010}}

@INPROCEEDINGS{Zhong2024ITSC,
  author={Zhong, Jiaru and Yu, Haibao and Zhu, Tianyi and Xu, Jiahui and Yang, Wenxian and Nie, Zaiqing and Sun, Chao},
  booktitle={2024 IEEE 27th International Conference on Intelligent Transportation Systems (ITSC)}, 
  title={Leveraging Temporal Contexts to Enhance Vehicle-Infrastructure Cooperative Perception}, 
  year={2024},
  volume={},
  number={},
  pages={915-922},
  doi={10.1109/ITSC58415.2024.10920140}}

@INPROCEEDINGS{8916930,
  author={Jin, Zhekai and Shao, Yifei and So, Minjoon and Sable, Carl and Shlayan, Neveen and Luchtenburg, Dirk Martin},
  booktitle={2019 IEEE Intelligent Transportation Systems Conference (ITSC)}, 
  title={A Multisensor Data Fusion Approach for Simultaneous Localization and Mapping}, 
  year={2019},
  volume={},
  number={},
  pages={1317-1322},
  doi={10.1109/ITSC.2019.8916930}}

@ARTICLE{Li9835036,
  author={Li, Yiming and Ma, Dekun and An, Ziyan and Wang, Zixun and Zhong, Yiqi and Chen, Siheng and Feng, Chen},
  journal={IEEE Robotics and Automation Letters}, 
  title={V2X-Sim: Multi-Agent Collaborative Perception Dataset and Benchmark for Autonomous Driving}, 
  year={2022},
  volume={7},
  number={4},
  pages={10914-10921},
  doi={10.1109/LRA.2022.3192802}
}

@article{hsu2018analysis,
  title={Analysis and modeling GPS NLOS effect in highly urbanized area},
  author={Hsu, Li-Ta},
  journal={GPS solutions},
  volume={22},
  number={1},
  pages={7},
  year={2018},
  publisher={Springer}
}

@ARTICLE{hf2022benchmark,
  author={Huang, Feng and Wen, Weisong and Zhang, Jiachen and Hsu, Li-Ta},
  journal={IEEE Intelligent Transportation Systems Magazine}, 
  title={{P}oint {W}ise or {F}eature {W}ise? {A} {B}enchmark {C}omparison of {P}ublicly {A}vailable {LiDAR} {O}dometry {A}lgorithms in {U}rban {C}anyons}, 
  year={2022},
  volume={14},
  number={6},
  pages={155-173},
  doi={10.1109/MITS.2021.3092731}}

@misc{deloitte_astri_2024,
  author       = {{Deloitte and ASTRI}},
  title        = {Hong Kong Connected \& Autonomous Vehicle (CAV) Development Study},
  howpublished = {\url{https://www2.deloitte.com/cn/zh/pages/technology/articles/hk-cav-development-study.html}},
  note         = {Accessed: April 18, 2024},
  year         = {2024},
  organization = {Deloitte}
}

@misc{yu2022dairv2x,
      title={DAIR-V2X: A Large-Scale Dataset for Vehicle-Infrastructure Cooperative 3D Object Detection}, 
      author={Haibao Yu and Yizhen Luo and Mao Shu and Yiyi Huo and Zebang Yang and Yifeng Shi and Zhenglong Guo and Hanyu Li and Xing Hu and Jirui Yuan and Zaiqing Nie},
      year={2022},
      eprint={2204.05575},
      archivePrefix={arXiv},
      primaryClass={cs.CV},
      url={https://arxiv.org/abs/2204.05575}, 
}

@ARTICLE{YIN2022M2DGR,
  author={Yin, Jie and Li, Ang and Li, Tao and Yu, Wenxian and Zou, Danping},
  journal={IEEE Robotics and Automation Letters}, 
  title={M2DGR: A Multi-Sensor and Multi-Scenario SLAM Dataset for Ground Robots}, 
  year={2022},
  volume={7},
  number={2},
  pages={2266-2273},
  doi={10.1109/LRA.2021.3138527}}

@INPROCEEDINGS{zimmer2024tumtrafv2x,
  author={Zimmer, Walter and Creß, Christian and Nguyen, Huu Tung and Knoll, Alois C.},
  booktitle={2023 IEEE 26th International Conference on Intelligent Transportation Systems (ITSC)}, 
  title={TUMTraf Intersection Dataset: All You Need for Urban 3D Camera-LiDAR Roadside Perception}, 
  year={2023},
  volume={},
  number={},
  pages={1030-1037},
  doi={10.1109/ITSC57777.2023.10422289}}

@INPROCEEDINGS{Radar2ThermalCalib,
  author={Zhang, Jun and Zhang, Shini and Peng, Guohao and Zhang, Haoyuan and Wang, Danwei},
  booktitle={2022 IEEE 25th International Conference on Intelligent Transportation Systems (ITSC)}, 
  title={3DRadar2ThermalCalib: Accurate Extrinsic Calibration between a 3D mmWave Radar and a Thermal Camera Using a Spherical-Trihedral}, 
  year={2022},
  volume={},
  number={},
  pages={2744-2749},
  doi={10.1109/ITSC55140.2022.9922522}}

@misc{xiang2024v2xreal,
      title={V2X-Real: a Large-Scale Dataset for Vehicle-to-Everything Cooperative Perception}, 
      author={Hao Xiang and Zhaoliang Zheng and Xin Xia and Runsheng Xu and Letian Gao and Zewei Zhou and Xu Han and Xinkai Ji and Mingxi Li and Zonglin Meng and Li Jin and Mingyue Lei and Zhaoyang Ma and Zihang He and Haoxuan Ma and Yunshuang Yuan and Yingqian Zhao and Jiaqi Ma},
      year={2024},
      eprint={2403.16034},
      archivePrefix={arXiv},
      primaryClass={cs.CV},
      url={https://arxiv.org/abs/2403.16034}, 
}

@INPROCEEDINGS{HUANG2023RSILIO,
  author={Huang, Feng and Chen, Hang and Urtay, Alpamys and Su, Dongzhe and Wen, Weisong and Hsu, Li-Ta},
  booktitle={2023 IEEE 26th International Conference on Intelligent Transportation Systems (ITSC)}, 
  title={Roadside Infrastructure assisted LiDAR/Inertial-based Mapping for Intelligent Vehicles in Urban Areas}, 
  year={2023},
  volume={},
  number={},
  pages={5831-5837},
  doi={10.1109/ITSC57777.2023.10422552}}

@ARTICLE{HUANG2024errormap,
  author={Huang, Feng and Wen, Weisong and Zhang, Guohao and Su, Dongzhe and Huang, Yulong},
  journal={IEEE Transactions on Instrumentation and Measurement}, 
  title={Continuous Error Map-Aided Adaptive Multisensor Integration for Connected Autonomous Vehicles in Urban Scenarios}, 
  year={2025},
  volume={74},
  number={},
  pages={1-13},
  doi={10.1109/TIM.2025.3573351}}

@INPROCEEDINGS{HUANG2025RSGLIO,
  author={Huang, Feng and Zhong, Yihan and Chen, Hang and Su, Dongzhe and Wu, Jin and Wen, Weisong and Hsu, Li-Ta},
  booktitle={2025 IEEE/RSJ International Conference on Intelligent Robots and Systems (IROS)}, 
  title={Roadside GNSS Aided Multi-Sensor Integrated System for Vehicle Positioning in Urban Areas}, 
  year={2025},
  volume={},
  number={},
  pages={},
  doi={}}

@article{KITTI,
  title={Vision meets robotics: The kitti dataset},
  author={Geiger, Andreas and Lenz, Philip and Stiller, Christoph and Urtasun, Raquel},
  journal={The International Journal of Robotics Research},
  volume={32},
  number={11},
  pages={1231--1237},
  year={2013},
  publisher={Sage Publications Sage UK: London, England}
}

@inproceedings{LidarCalibrMars2022,
  title={Robust real-time lidar-inertial initialization},
  author={Zhu, Fangcheng and Ren, Yunfan and Zhang, Fu},
  booktitle={2022 IEEE/RSJ International Conference on Intelligent Robots and Systems (IROS)},
  pages={3948--3955},
  year={2022},
  organization={IEEE}
}

@article{VINS-MONO,
  title={Vins-mono: A robust and versatile monocular visual-inertial state estimator},
  author={Qin, Tong and Li, Peiliang and Shen, Shaojie},
  journal={IEEE Transactions on Robotics},
  volume={34},
  number={4},
  pages={1004--1020},
  year={2018},
  publisher={IEEE}
}

@inproceedings{zhang2023ntu4dradlm,
  title={Ntu4dradlm: 4d radar-centric multi-modal dataset for localization and mapping},
  author={Zhang, Jun and Zhuge, Huayang and Liu, Yiyao and Peng, Guohao and Wu, Zhenyu and Zhang, Haoyuan and Lyu, Qiyang and Li, Heshan and Zhao, Chunyang and Kircali, Dogan and others},
  booktitle={2023 IEEE 26th International Conference on Intelligent Transportation Systems (ITSC)},
  pages={4291--4296},
  year={2023},
  organization={IEEE}
}

@inproceedings{Lio-sam,
  title={Lio-sam: Tightly-coupled lidar inertial odometry via smoothing and mapping},
  author={Shan, Tixiao and Englot, Brendan and Meyers, Drew and Wang, Wei and Ratti, Carlo and Rus, Daniela},
  booktitle={2020 IEEE/RSJ international conference on intelligent robots and systems (IROS)},
  pages={5135--5142},
  year={2020},
  organization={IEEE}
}

@article{Fast-lio2,
  title={Fast-lio2: Fast direct lidar-inertial odometry},
  author={Xu, Wei and Cai, Yixi and He, Dongjiao and Lin, Jiarong and Zhang, Fu},
  journal={IEEE Transactions on Robotics},
  volume={38},
  number={4},
  pages={2053--2073},
  year={2022},
  publisher={IEEE}
}

@ARTICLE{FAST-LIVO2,
  author={Zheng, Chunran and Xu, Wei and Zou, Zuhao and Hua, Tong and Yuan, Chongjian and He, Dongjiao and Zhou, Bingyang and Liu, Zheng and Lin, Jiarong and Zhu, Fangcheng and Ren, Yunfan and Wang, Rong and Meng, Fanle and Zhang, Fu},
  journal={IEEE Transactions on Robotics}, 
  title={FAST-LIVO2: Fast, Direct LiDAR–Inertial–Visual Odometry}, 
  year={2025},
  volume={41},
  number={},
  pages={326-346},
  doi={10.1109/TRO.2024.3502198}}

@article{GigE,
  title={GiGE Vision},
  author={Automated Imaging Association and others},
  journal={GigE Vision Specification version},
  volume={1},
  number={0},
  pages={1},
  year={2006}
}

@inproceedings{PTP,
  title={Software and hardware prototypes of the IEEE 1588 precision time protocol on wireless LAN},
  author={Kannisto, Juha and Vanhatupa, Timo and Hannikainen, M and Hamalainen, TD},
  booktitle={2005 14th IEEE Workshop on Local \& Metropolitan Area Networks},
  pages={6--pp},
  year={2005},
  organization={IEEE}
}

@ARTICLE{icp1992,
  author={Besl, P.J. and McKay, Neil D.},
  journal={IEEE Transactions on Pattern Analysis and Machine Intelligence}, 
  title={A method for registration of 3-D shapes}, 
  year={1992},
  volume={14},
  number={2},
  pages={239-256},
  doi={10.1109/34.121791}}

@phdthesis{Martin2009,
author = {Magnusson, Martin},
school = {Örebro University},
year = {2009},
month = {12},
pages = {},
title = {The Three-Dimensional Normal-Distributions Transform --- an Efficient Representation for Registration, Surface Analysis, and Loop Detection}
}

@article{langley1995nmea,
  title={Nmea 0183: A gps receiver},
  author={Langley, Richard},
  journal={GPS world},
  volume={6},
  number={7},
  pages={54--57},
  year={1995}
}

@inproceedings{NovAtel,
  title={Architecture and system performance of SPAN-NovAtel's GPS/INS solution},
  author={Kennedy, Sandy and Hamilton, Jason and Martell, Hugh},
  booktitle={Proceedings of IEEE/ION PLANS 2006},
  pages={266--274},
  year={2006}
}

@inproceedings{nguyen2024mcd,
  title={Mcd: Diverse large-scale multi-campus dataset for robot perception},
  author={Nguyen, Thien-Minh and Yuan, Shenghai and Nguyen, Thien Hoang and Yin, Pengyu and Cao, Haozhi and Xie, Lihua and Wozniak, Maciej and Jensfelt, Patric and Thiel, Marko and Ziegenbein, Justin and others},
  booktitle={Proceedings of the IEEE/CVF Conference on Computer Vision and Pattern Recognition},
  pages={22304--22313},
  year={2024}
}

@inproceedings{furgale2013unified,
  title={Unified temporal and spatial calibration for multi-sensor systems},
  author={Furgale, Paul and Rehder, Joern and Siegwart, Roland},
  booktitle={2013 IEEE/RSJ International Conference on Intelligent Robots and Systems},
  pages={1280--1286},
  year={2013},
  organization={IEEE}
}

@inproceedings{google2022gsdc,
   author = {Fu, Michael and Khider, Mohammed and van Diggelen, Frank},
   title = {Summary and Legacy of the Smartphone Decimeter Challenge (SDC) 2022},
   booktitle = {Proceedings of the 36th International Technical Meeting of the Satellite Division of The Institute of Navigation (ION GNSS+ 2023)},
   address = {Denver, Colorado},
   pages = {2301 - 2320},
   DOI = {https://doi.org/10.33012/2022.18379},
   year = {2022},
   month = {September},
}

@ARTICLE{DCL-SLAM2024,
  author={Zhong, Shipeng and Qi, Yuhua and Chen, Zhiqiang and Wu, Jin and Chen, Hongbo and Liu, Ming},
  journal={IEEE Sensors Journal}, 
  title={DCL-SLAM: A Distributed Collaborative LiDAR SLAM Framework for a Robotic Swarm}, 
  year={2024},
  volume={24},
  number={4},
  pages={4786-4797},
  doi={10.1109/JSEN.2023.3345541}}

@INPROCEEDINGS{Schmuck2017,
  author={Schmuck, Patrik and Chli, Margarita},
  booktitle={2017 IEEE International Conference on Robotics and Automation (ICRA)}, 
  title={Multi-UAV collaborative monocular SLAM}, 
  year={2017},
  volume={},
  number={},
  pages={3863-3870},
  doi={10.1109/ICRA.2017.7989445}}

@article{Schmuck2019,
author = {Schmuck, Patrik and Chli, Margarita},
title = {CCM-SLAM: Robust and efficient centralized collaborative monocular simultaneous localization and mapping for robotic teams},
journal = {Journal of Field Robotics},
volume = {36},
number = {4},
pages = {763-781},
doi = {https://doi.org/10.1002/rob.21854},
url = {https://onlinelibrary.wiley.com/doi/abs/10.1002/rob.21854},
eprint = {https://onlinelibrary.wiley.com/doi/pdf/10.1002/rob.21854},
year = {2019}
}

\end{document}